\documentclass[10pt,twocolumn,letterpaper]{article}

\usepackage{cvpr}              

\newcommand{\myparagraph}[1]{{\noindent\bf #1}}
\usepackage[misc]{ifsym} 

\makeatletter
\newcommand\blfootnote[1]{%
  \begingroup
  \renewcommand\thefootnote{}%
  \renewcommand\@makefntext[1]{\noindent ##1}
  \footnote{#1}%
  \addtocounter{footnote}{-1}%
  \endgroup
}
\makeatother

\newcommand{\nickname}{\textsc{Video4Spatial}\xspace}

\definecolor{cvprblue}{rgb}{0.21,0.49,0.74}
\usepackage[pagebackref,breaklinks,colorlinks,allcolors=cvprblue]{hyperref}
\usepackage{caption}
\usepackage{xcolor}
\usepackage[table]{xcolor} 
\usepackage[svgnames]{xcolor}
\usepackage{booktabs}
\usepackage{comment}
\usepackage[most]{tcolorbox}
\usepackage{enumitem}
\usepackage{lipsum}

\def\paperID{7487} 
\def\confName{CVPR}
\def\confYear{2026}

\title{{\nickname: Towards Visuospatial Intelligence \\with Context-Guided Video Generation}} 

\author{%
  Zeqi Xiao$^{1,2*}$\quad  Yiwei Zhao$^{1\textrm{\Letter}\dagger}$\quad  Lingxiao Li$^{1}$\quad  Yushi Lan$^{3}$, \\
  Ning Yu$^{4}$\quad  Rahul Garg$^{1}$\quad  Roshni Cooper$^{1}$\quad Mohammad H. Taghavi$^{1}$\quad Xingang Pan$^{2\textrm{\Letter}}$ \vspace{3pt} \\
  $^{1}$Netflix\quad
  $^{2}$Nanyang Technological University, \\
  $^{3}$University of Oxford \quad
  $^{4}$Netflix Eyeline Studios \\
  \footnotesize \texttt{\{zeqi001, xingang.pan\}@ntu.edu.sg}\\
  \footnotesize \texttt{\{yiweiz, lingxiaol, rahulgarg, mtaghavi\}@netflix.com} \\
  \footnotesize \texttt{yushi.lan@eng.ox.ac.uk}\quad
  \texttt{ning.yu@scanlinevfx.com}
}

\newif\ifdraft
\drafttrue

\ifdraft
    \newcommand{\yiwei}[1]{\textbf{\textcolor{red}{ZY: #1}}}
    \newcommand{\zeqi}[1]{\textbf{\textcolor{blue}{ZQ: #1}}}
    \newcommand{\lingxiao}[1]{\textbf{\textcolor{purple}{LL: #1}}}
    \newcommand{\rahul}[1]{\textbf{\textcolor{cyan}{RG: #1}}}
    \newcommand{\ning}[1]{\textbf{\textcolor{DarkGreen}{Ning: #1}}}
    \newcommand{\xingang}[1]{\textbf{\textcolor{brown}{XG: #1}}}
    
\else
    \newcommand{\yiwei}[1]{}
    \newcommand{\zeqi}[1]{}
    \newcommand{\lingxiao}[1]{}
    \newcommand{\rahul}[1]{}
    \newcommand{\ning}[1]{}
    \newcommand{\xingang}[1]{}
\fi

\usepackage{soul}

\begin{document}

\twocolumn[{
    \renewcommand\twocolumn[1][]{#1}
    \maketitle
    \centering
    \includegraphics[width=\linewidth]{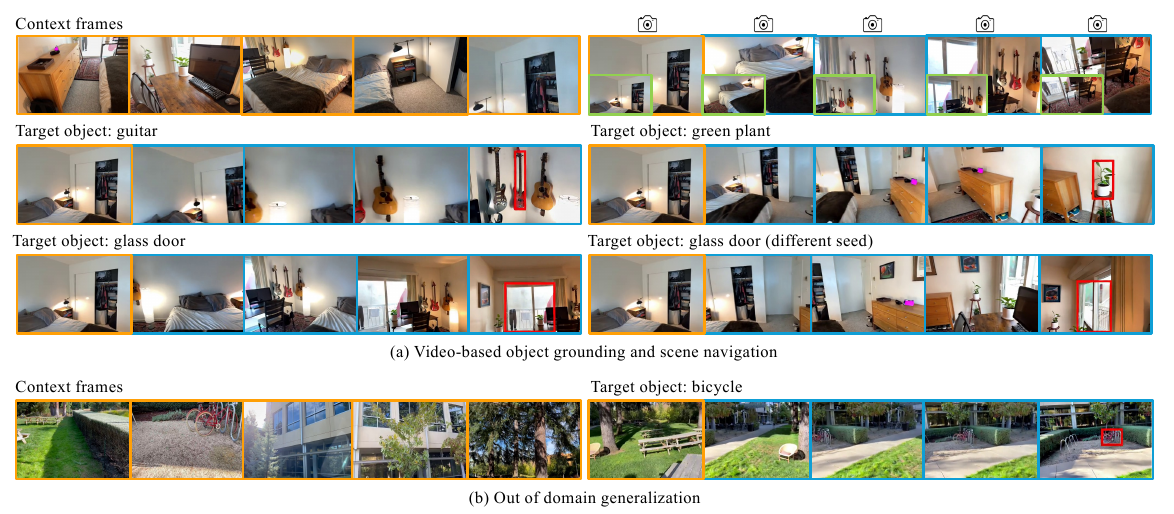}
        \captionof{figure}{Our method generates videos that fulfill instructional spatial tasks while remaining geometrically consistent with the provided video context. (a) We demonstrate two tasks: \textbf{video-based object grounding}, where the model follows text instructions (the second and the third row) to navigate and locate a target object (\eg, ``a green plant" or ``a guitar" ).
        Navigation paths and final locations are implicitly planned with diversity by model during generation to achieve the same goal (third row); and \textbf{scene navigation} (top-right), where the model generates a video that adheres to a specified camera trajectory (denoted by camera logos) and aligns with ground truth videos (\textcolor{green!60!black}{green boxes})
        (b) Although trained on indoor datasets, our model generalizes well to out-of-domain scenarios, such as an outdoor park.
        \textcolor{orange}{Orange boxes} denote context frames. \textcolor{cyan}{Blue boxes}
        denote generated frames.  Please note that \textcolor{red}{red bounding boxes} are generated by the video model itself. Frames are subsampled for visualization. Project page at \href{https://xizaoqu.github.io/video4spatial/}{https://xizaoqu.github.io/video4spatial}.}
         \vspace{20pt}
    \label{fig:teaser}
}]

\blfootnote{* Work done during the internship at Netflix. \\ \textrm{\Letter} Corresponding Author. $\dagger$ Project Lead.}

\begin{abstract}
We investigate whether video generative models can exhibit visuospatial intelligence, a capability central to human cognition, using only visual data. To this end, we present \nickname, a framework showing that video diffusion models conditioned solely on video‑based scene context can perform complex spatial tasks. We validate on two tasks: scene navigation - following camera‑pose instructions while remaining consistent with 3D geometry of the scene, and object grounding - which requires semantic localization, instruction following, and planning. 
Both tasks use video‑only inputs, without auxiliary modalities such as depth or poses. With simple yet effective design choices in the framework and data curation, \nickname demonstrates strong spatial understanding from video context: it plans navigation and grounds target objects end‑to‑end, follows camera‑pose instructions while maintaining spatial consistency, and generalizes to long contexts and out‑of‑domain environments. 
Taken together, these results advance video generative models toward general visuospatial reasoning.
\end{abstract}    
\section{Introduction}
\label{sec:intro}

Humans excel at remembering, understanding, and acting within spatial environments—a hallmark of high-level spatial reasoning capability. A long line of research has sought to endow neural networks with similar capabilities. Recently, rapidly advancing video generative models \cite{veo3, sora2, wan2025wan} have begun to address this challenge, exhibiting general-purpose perceptual and reasoning abilities \cite{wiedemer2025video}, echoing the trajectory of Large Language Models~\cite{gpt4o,claude35}. Because video generation is inherently sequential and temporally coherent, these models are particularly well-suited to visuospatial-related tasks: by rolling out future frames that preserve scene geometry and appearance, they can simulate plausible futures. For example, in maze navigation, a model can “imagine” progress by generating frames that maintain the maze’s topology while ensuring smooth transitions~\cite{wiedemer2025video, tong2025thinking}.

Humans primarily perceive spatial information visually. We aim to endow models with the same capability by using video as a simple, general, and scalable modality. However, learning spatial understanding solely from RGB video is challenging, and many recent methods still rely on auxiliary signals (\eg, depth maps, camera poses, point clouds) for supervision or augmentation~\cite{zhu2024llava, xu2024vlm, wu2025spatial,wei2025streamvln,yu2025trajectorycrafter}. In this paper, we present \nickname and show that \emph{video generative models that rely solely on video context of a spatial environment can exhibit strong spatial intelligence}. Our framework is intentionally simple: a standard video diffusion architecture trained only with the diffusion objective. The inputs are a video context (several frames from the same environment) and instructions; the output is a video that completes the instructed spatial task while maintaining scene geometry and temporal coherence.

The core spatial abilities our framework aims to achieve include (i) implicitly inferring 3D structure ~\cite{zhou2017unsupervised,jiang2025geo4d}, (ii) controlling viewpoint with geometric consistency (scale, occlusion, layout)~\cite{wang2024motionctrl,he2024cameractrl}, (iii) understanding semantic layout~\cite{li2024llava}, (iv) planning toward goals~\cite{anderson2018vision}, and (v) maintaining long-horizon temporal coherence. 
We introduce two tasks to probe these abilities: \textit{video-based scene navigation and object grounding}. 
In scene navigation, the model follows camera-pose instructions and produces a video whose trajectory is consistent with both the instructions and the scene geometry. In object grounding, a text instruction asks the model to reach and visibly localize a target object, thereby testing semantic layout understanding and plausible planning (Fig.~\ref{fig:teaser}). 
Together, the tasks evaluate whether \nickname can infer and act on scene structure using video alone.

Beyond the base architecture, our central design choice is how to model conditions on \emph{video context}. In line with DiT~\cite{peebles2023scalable}, we process context and target frames through the same transformer stack, while setting the diffusion timestep of context frames to $t=0$. Inspired by History Guidance~\cite{song2025history}, we extend classifier‑free guidance~\cite{ho2022classifier} to video context, which we find significantly improves contextual coherence. 
To reduce redundancy in continuous footage in context, we subsample non‑contiguous frames during training and inference and apply non‑contiguous RoPE~\cite{su2024roformer} over the corresponding subsampled indices. 
This index sparsity also enables extrapolation to much longer contexts at inference. 
For visuospatial tasks, we find that explicit reasoning patterns~\cite{gandhi2025cognitive} help task completion; accordingly, we train the model to predict videos with visual bounding boxes to strengthen object grounding.

Since no large public datasets are purpose-built for these spatial task settings, we repurpose indoor scanning datasets-ScanNet++~\cite{yeshwanth2023scannet++} and ARKitScenes~\cite{baruch2021arkitscenes}, which provide long trajectories through diverse rooms and enable sampling of context–target video clip pairs. We use an off-the-shelf vision-language model (VLM)~\cite{qwen3vl} to (i) identify clips that end with a centered object and (ii) generate natural-language grounding instructions; for navigation, we convert the annotated camera poses into pose-following instructions. 

Experimental results demonstrate that the proposed framework effectively addresses challenging visuospatial tasks. 
From video context alone, the model acquires scene-level geometry and semantics, and achieves substantially strong performance on object grounding and scene navigation. 
Trained with short contexts, \nickname extrapolates at inference to substantially longer temporal windows, further improving performance. Meanwhile, though being trained on indoor scenes, it generalizes to outdoor environments and to object categories unseen during training.


In summary, we: (1) introduce \nickname, a simple yet effective video generation framework that relies solely on video‑based scene context to perform visuospatial tasks; (2) highlight key design choices—joint classifier‑free guidance over context and instruction, auxiliary bounding boxes as a reasoning prior to improve grounding accuracy, and non‑continuous context sampling for efficient context understanding; and (3) instantiate and evaluate two spatial reasoning tasks: scene navigation and object grounding, which require consistency with 3D geometry and semantic layout without auxiliary modalities (\eg, depth, point clouds). These results advance video‑only spatial reasoning, and we hope our work will inspire future research on leveraging video generation models for visuospatial intelligence.

\section{Related Work}

\myparagraph{Generative Video Models: from Renderer to Reasoner.}
Building on advances in diffusion models~\cite{song2020score, peebles2023scalable, chen2025diffusion}, video generation has progressed rapidly~\cite{wang2023modelscope, wang2023lavie, chen2023videocrafter1, guo2023animatediff, sora, jin2024pyramidal, yin2024slow}. Video diffusion models can function as high-fidelity, controllable renderers, with generation steered by control signals such as camera control~\cite{wang2024motionctrl, he2024cameractrl, xiao2024trajectory, bai2025recammaster,bar2024navigationworldmodels}, sketches~\cite{wang2024videocomposer}, depth maps~\cite{wang2024videocomposer}, trajectories~\cite{yin2023dragnuwa, teng2023drag, deng2023dragvideo, fu20243dtrajmaster,fu2025learning}, point cloud ~\cite{yu2024viewcrafter, yu2025trajectorycrafter, ren2025gen3c} and human poses~\cite{mimicmotion2024, zhu2024champ}.

Powered by the rapid growth of web-scale training, video models~\cite{sora2, veo3, Kling, agarwal2025cosmos} have begun to exhibit surprising world priors~\cite{li2025worldmodelbench, zheng2025vbench, meng2024towards,genie3} beyond visual fidelity, like physical awareness~\cite{meng2024towards} and reasoning ability~\cite{wiedemer2025video}. In particular, \citet{wiedemer2025video} demonstrate the potential of generative video models on reasoning tasks such as maze solving and robot navigation, marking a transition from mere rendering to genuine reasoning. Our work focuses on video model's visuospatial reasoning capability with visual context.

\myparagraph{Video-to-Video Generation.}
Beyond generating videos solely from text or other multimodal controls, video-to-video generation~\cite{mallya2020world, wang2019few, wang2018video} conditions on an input video to produce a new video. It has been explored across tasks such as video editing~\cite{liu2024video, ouyang2024i2vedit, yang2023rerender}, video outpainting~\cite{chen2024follow}, video super-resolution~\cite{xu2025videogigagan, zhou2024upscale, wang2025seedvr}, and camera control~\cite{bai2025recammaster, xiao2024trajectory, gu2025diffusion}. These settings typically adopt frame-to-frame transformations, which constrain the output’s content and duration. A more general paradigm treats the input video as context while allowing free-form generation, requiring stronger long-range modeling and attention. Recent efforts in this direction include multi-shot movie generation~\cite{guo2025long, cai2025mixture, xiao2025captain} and autoregressive video generation~\cite{huang2025self, zhang2025packing}. However, how to effectively leverage scene-level video context for spatial tasks remains underexplored. 

\myparagraph{Visuospatial Intelligence.}
Visuospatial Intelligence (VSI) ~\cite{yang2025thinking} is the ability to perceive~\cite{wang2025vggt,wang2024dust3r}, represent~\cite{mildenhall2021nerf,kerbl20233d}, and act~\cite{chen2025gleam,wei2025streamvln,anderson2018vision} spatially from visual input. Prior works can be roughly categorized by output modality. Many studies probe visuospatial intelligence with vision–language models (VLMs)~\cite{li2024llava,zhang2024long,yin2025spatial}, where the output is text. Conditioning language models on visuals alone often falls short, so several methods add explicit 3D signals (\eg, point clouds, depth) to the context window~\cite{zhu2024llava, xu2024vlm, wu2025spatial}. Recent work also brings visual generation into the reasoning loop and shows promise on spatial tasks~\cite{yang2025machine, yang2025mindjourney}.  Alternatively, many works study VSI through video generation~\cite{jiang2025geo4d,chen2025video}, where the output is video. For spatially consistent video generation, some reconstruct external 3D memories (\eg, point clouds)~\cite{wu2025video, li2025vmem} extracted by 3D visual foundation models~\cite{wang2025vggt,wang2024dust3r}. Other approaches rely solely on visual input but still require camera-pose annotations~\cite{xiao2025worldmem, yu2025context, chen2025learning,zhou2025stable, zhang2025test}. In contrast, we show that conditioning video generative models solely on raw video sequences suffices for spatial understanding, enabling an end-to-end, scalable approach without explicit 3D signals like depth or pose.



\section{Methods}

This section formalizes video generation as spatial reasoning (Section~\ref{sec:spatial_reasoner}) and describes the evaluation tasks (Section~\ref{sec:tasks}) and key design choices (Section~\ref{sec:designs}).

\subsection{Preliminaries}
\myparagraph{Video Diffusion Models (VDMs).}
Diffusion models~\cite{sohl2015deep,ho2020denoising,song2020score} learn a score function
\(s_\theta(\mathbf{x}, t, c) \approx \nabla_{\mathbf{x}}\log p_t(\mathbf{x}\mid c)\),
where \(p_t\) is the data distribution smoothed by Gaussian noise at level \(t\), and \(c\) denotes conditioning.
At inference, a sample from the standard Gaussian is iteratively denoised using \(s_\theta\) to recover the data distribution.
VDMs~\cite{ho2022video} produce high-fidelity, temporally consistent videos, benefiting from transformer backbones~\cite{vaswani2017attention,peebles2023scalable} and latent diffusion~\cite{rombach2022high}.
Typical conditioning includes the first frame~\cite{blattmann2023stable,yang2024cogvideox} and/or a text instruction~\cite{ho2022video,wang2023modelscope}.
History-conditioned video generation~\cite{song2025history} shows that \(c\) can include preceding video frames treated as part of the generation sequence inside the transformer.
We adopt this architectural template and extend \(c\) to \emph{general video context} (Section~\ref{sec:spatial_reasoner}).

\myparagraph{Classifier-Free Guidance (CFG).}
CFG~\cite{ho2022classifier} improves conditional sampling by learning with randomly dropped conditioning (\(c=\varnothing\)) and, at sampling, using
\begin{equation}
\label{eq:cfg}
s_\theta(\mathbf{x},t,\varnothing) \;+\;
\omega\!\left(s_\theta(\mathbf{x},t,c)\;-\;s_\theta(\mathbf{x},t,\varnothing)\right),
\end{equation}
where \(\omega\!\ge\!1\) is the guidance scale.

\myparagraph{Rotary Positional Embeddings (RoPE).}
RoPE~\cite{su2024roformer} injects relative spatial–temporal position by rotating queries and keys, enabling attention by displacement rather than absolute indices.
In video diffusion, RoPE improves alignment across frames and stabilizes motion~\cite{wan2025wan,opensora}.
In practice, attention heads can be partitioned over spatial/temporal axes with per-axis frequency scaling to match resolution and clip length.



\begin{figure}[t]
    \centering
    \includegraphics[width=0.9\linewidth]{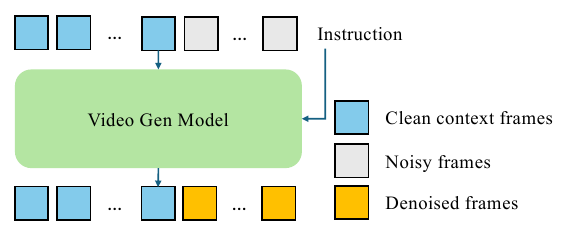}
    \caption{We frame video diffusion models as spatial reasoners, treating context and target frames equally in the model architecture except that context frames are noise-free.}
    \label{fig:framework}
\end{figure}

\subsection{Video Generation Models as Spatial Reasoners}\label{sec:spatial_reasoner}

Prior work often frames video generative models as \emph{neural renderers}~\cite{gu2025diffusion, ren2025gen3c}, prioritizing visual fidelity while overlooking the reasoning potential afforded by large-scale pretraining on web-scale corpora rich in structural spatial knowledge.
We instead cast video generation as \emph{spatial reasoning}: given video-only context, the model produces coherent, goal-directed outcomes following instructions. 

Concretely, given a context video $\mathbf{x}_{\text{ctx}}$ ( frames from the observed environment) and an instruction $\mathbf{g}$ (textual, visual, or other structured forms), the model synthesizes 
\begin{equation}\label{eqn:task_format}
    \mathbf{x}_{\text{out}} \sim p_\theta\!\big(\cdot \mid \mathbf{x}_{\text{ctx}}, \mathbf{g}\big) \approx p(\cdot \mid \mathbf{x}_{\text{ctx}},\mathbf{g}),
\end{equation}
where $p_\theta$ is the model distribution and $p$ the true posterior over videos consistent with $\mathbf{x}_{\text{ctx}}$ and $\mathbf{g}$.
We implement context condition by concatenating \(\mathbf{x}_{\text{ctx}}\) with the noisy target frames along the temporal axis (Fig.~\ref{fig:framework}); 
context frames are fixed at noise-free ($t=0)$ during training and inference, while the target frames share the same diffusion noise level.
The instruction tokens (text or structured signals such as poses) can be injected via cross‑attention~\cite{wan2025wan}, learned embedding addition~\cite{he2024cameractrl}, or token-wise concatenation~\cite{esser2024scaling, bai2025recammaster}.

\subsection{Solving Spatial Tasks with Videos}\label{sec:tasks}

We explore video model's spatial reasoning capability in realistic environments (homes, offices, factories) using the following two tasks:
\emph{object grounding} (text instruction) and \emph{scene navigation} (pose instruction).
Both tasks assess how well a model can infer 3D structure, maintain contextual consistency, and generate coherent motion that aligns with a given instruction. 
Both tasks use the same format described in Eq.~\ref{eqn:task_format}; unless stated otherwise, the first target frame is also provided as part of $\mathbf{x}_{\text{ctx}}$.

\myparagraph{Video-based object grounding.}
Given a context video and a natural-language instruction specifying a target object, the model generates a sequence that moves from the initial viewpoint and ends with the target prominently localized in the final frame. We evaluate this task through outputs' 3D geometric consistency with the context and accuracy to find the object, details in Sec.~\ref{sec:eval_grounding}.

\myparagraph{Video-based scene navigation.}
Given a context video and a sequence of \emph{egocentric} 6-DoF pose waypoints \(\{(\Delta\mathbf{t}_k,\Delta\mathbf{r}_k)\}_k\) (\(\Delta\mathbf{t}\in\mathbb{R}^3\) in meters; \(\Delta\mathbf{r}\) as yaw–pitch–roll in radians), the model synthesizes novel viewpoints along the specified trajectory (details in Sec.\ref{sec:eval_scene_nav}).
While related to novel view synthesis, we focus on navigation via continuous video generation without explicit 3D information (\eg, camera pose or depth of the context, or reconstructed geometry).

\subsection{Key Design Choices}\label{sec:designs}

\myparagraph{Joint CFG~\cite{ho2022classifier}.}
Inspired by History Guidance~\cite{song2025history}, we use the following CFG formulation on the joint conditioning of instruction $\mathbf{g}$ and $\mathbf{x}_{\text{ctx}}$:
\begin{flalign*}
s_\theta(\mathbf{x},t,\varnothing,\mathbf{z}_{\text{ctx}}) 
+ \omega(s_\theta(\mathbf{x},t,\mathbf{g},\mathbf{x}_{\text{ctx}})-s_\theta(\mathbf{x},t,\varnothing,\mathbf{z}_{\text{ctx}})), 
\end{flalign*}
where $\mathbf{z_{\text{ctx}}} \sim\mathcal{N}\big(0,\,\mathbf{I}\big)$ is pure noise. 
As shown in Table~\ref{tab:grounding_eval}, such joint CFG significantly improves the quality and consistency of the output.

\begin{figure}[t]
    \centering
    \includegraphics[width=\linewidth]{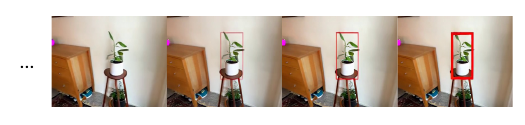}
    \vspace{-16pt}
    \caption{\textbf{Auxiliary bounding box.} The model is trained to generate a bbox (pixel values) located at the target object in the final frames as an explicit reasoning pattern~\cite{gandhi2025cognitive}. This auxiliary output improves object grounding accuracy.}
    \vspace{-12pt}
    \label{fig:bbox}
\end{figure}

\myparagraph{Auxiliary bounding box for object grounding.}
Injecting explicit reasoning patterns have been shown to improve language models' capabilities~\cite{gandhi2025cognitive}.
By exploiting the fact that video is a flexible medium, for the object grounding task, we annotate and augment the training videos at the end with a red bounding box (bbox) centered at the target object (Fig.~\ref{fig:bbox}).
This encourages the model to draw a bbox, reinforcing the grounding objective, leading to substantially improved accuracy (Table.~\ref{tab:grounding_eval}).

\begin{figure}[t]
    \centering
    \includegraphics[width=\linewidth]{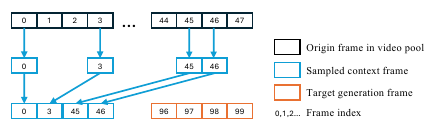}
    \vspace{-12pt}
    \caption{\textbf{RoPE indexing for non-contiguous context}. Context frames(\textcolor{cyan}{blue}) are sampled sparsely to reduce redundancy but retain their original temporal indices. Target generation frames (\textcolor{orange}{orange}) are given new indices with a fixed offset.}
    \vspace{-12pt}
    \label{fig:contextrope}
\end{figure}

\myparagraph{Non‑contiguous context sampling.}
A context video of contiguous frames exhibits substantial redundancy. To increase information content while keeping computation tractable, we train by sampling non‑contiguous context frames from source videos. Crucially, we apply RoPE to these sparse frames using their original source indices (Fig.~\ref{fig:contextrope}). This encourages the model to learn temporal coherence and improves robustness when extrapolating to longer context lengths at inference.

Additionally, we must also distinguish context frames from the target frames being generated, especially to prevent information leakage when both are sampled from the same source video. Therefore, inspired by ~\cite{tan2025ominicontrol}, we set the RoPE indices for the generated frames to begin after the final context frame's index, separated by a fixed interval of 50.
It ensures a clear separation between the context and the target, and mitigates potential information leakage.

\section{Experiments}

\begin{table*}[t]
\centering
\caption{\textbf{Evaluation on video-conditioned object grounding.} Spatial Distance (SD) evaluates 3D geometric consistency with the context. Instruction Following Rate (IF) measures the model’s ability to follow instructions. IF (SD$<\delta$) measures the overall ability to ground the target object while maintaining spatial geometry consistency. Imaging Quality (IQ) and Dynamic Degree (DD) evaluate overall video quality. We resize all results to the same resolution for evaluation. 
}
\label{tab:grounding_eval}
\small
\resizebox{0.87\textwidth}{!}{
\begin{tabular}{l | c c c | c c c c | c c}
\toprule
Methods  & Resolution & \# Context & \# Generation & SD $\downarrow$ & IF $\uparrow$ & \textbf{IF (SD$<$0.2)} $\uparrow$ & \textbf{IF (SD$<$0.1)} $\uparrow$ & IQ $\uparrow$ & DD $\uparrow$ \\
\midrule
GT* & 416$\times$256 & - & 161 & 0.0739 & 1.0 & 1.0 & 0.8260 & 0.6539 & 0.9783\\
\midrule
Wan2.2-5B~\cite{wan2025wan} & 1280$\times$704 & 1 & 121 & 0.5341 & \underline{0.9439} & 0.2242& 0.0934 & 0.6101 & 0.9277\\
Veo3~\cite{veo3} & 1280$\times$720 & 1 & 192 & \underline{0.2211} & \textbf{0.9532} & \underline{0.4599} & \underline{0.3821} & \textbf{0.7056} & 0.7487 \\
FramePack~\cite{zhang2025packing} & 704$\times$544 & 104+1 & 180 & 0.3672 & 0.3 & 0.1 & 0.1 &  0.6315 & \underline{0.9678} \\
Ours & 416$\times$256 & 336+1 & 161+20 & \textbf{0.1099} & 0.7327 & \textbf{0.6486} & \textbf{0.4941} & \underline{0.6435} & \textbf{0.9859} \\
\midrule
Ours w/o pretraining & 416$\times$256 & 336+1 & 161+20 & 0.4317 & 0.4990 & 0.2186 & 0.1214 & 0.5845 & 0.9803 \\
Ours w/ only first frame & 416$\times$256 & 1 & 161+20 & 0.4053 & 0.7700 & 0.2616 & 0.1084 & 0.6219 & 0.9842 \\
Ours w/o CFG & 416$\times$256 & 336+1 & 161+20 & 0.3890 & 0.4042 & 0.2031 & 0.1113 & 0.5299 & 0.9985\\
Ours w/o context CFG & 416$\times$256 & 336+1 & 161+20 & 0.8294 & 0.8841 & 0.0373 & 0.0112 & 0.4909 & 0.9937 \\
Ours w/o auxiliary bbox & 416$\times$256 & 336+1 & 161 & 0.1102 & 0.6191 & 0.5401 & 0.3551 & 0.6456 & 0.9883 \\
Ours w/ vanilla RoPE & 416$\times$256 & 336+1 & 161+20 & 0.2079 & 0.7383 & 0.4719 & 0.2570& 0.6239 & 0.8906\\
\bottomrule
\end{tabular}
}
\vspace{-6pt}
\end{table*}

\begin{figure*}[t]
    \centering
    \includegraphics[width=\linewidth]{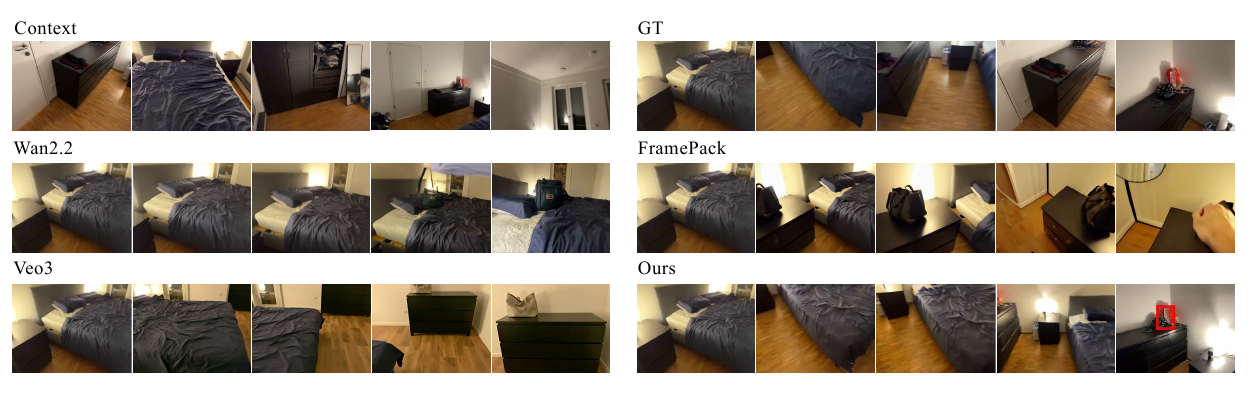}
    \vspace{-12pt}
    \caption{\textbf{Qualitative comparison across methods.} The instruction is: ``The camera moves smoothly through a bedroom. Finally, it focuses on \textbf{a bag} on top of the dresser in the center of the frame.'' Our method faithfully grounds the object present in the context, whereas other methods hallucinate the target object. Notably, the path to locate the target from our generated result is different from the GT. For the best experience, see the supplementary videos.}
    \label{fig:qualitative}
    \vspace{-16pt}
\end{figure*}

\myparagraph{Implementation Details.}
We fine‑tune the attention layers of video generation model Wan2.2~\cite{wan2025wan} with a flow‑matching objective~\cite{lipman2022flow}. We use a 1e-4 learning rate and a batch size of 160 and train for 20K steps for object grounding and 10K steps for scene navigation. Unless stated otherwise, we use 337 context frames and generate 161 frames. Conditioning is provided via (i) text instructions injected through cross‑attention or (ii) camera poses encoded with Plücker coordinates~\cite{sitzmann2021light} and added to the frame tokens. See the supplementary materials for additional details.

\begin{figure}[t]
    \centering
    \includegraphics[width=\linewidth]{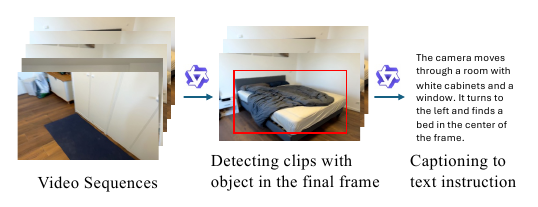}
    \vspace{-18pt}
    \caption{Object grounding data collection pipeline.}
    \label{fig:data_collection}
\end{figure}

\myparagraph{Datasets and Training.} Datasets are curated from ScanNet++~\cite{yeshwanth2023scannet++} and ARKitScenes~\cite{baruch2021arkitscenes}. For object grounding, we use Qwen3-VL~\cite{qwen3vl} to identify video clips with an object centered in the last frame and generate text instructions, resulting in 400K text-video pairs (Fig.~\ref{fig:data_collection}). 
For scene exploration, we directly use ScanNet++ with 1K videos and camera pose annotations and randomly sample short clips for training. For both tasks, we evaluate on randomly selected 18 scenes from ScanNet++ that are not included in the training set.

\begin{figure}[t]
    \centering
    \includegraphics[width=\linewidth]{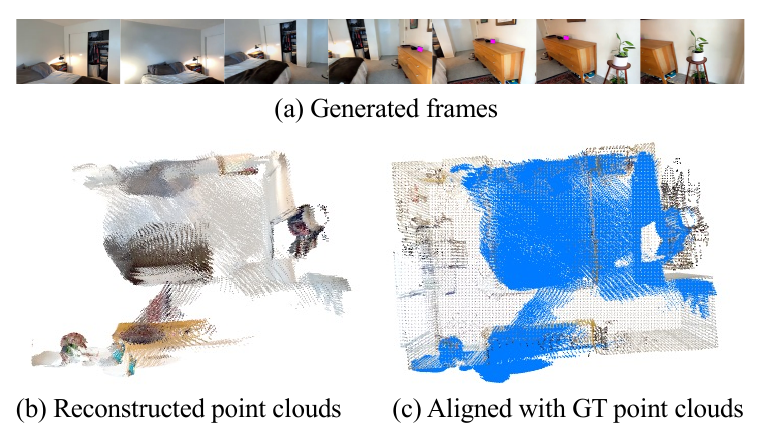}
    \caption{\textbf{Visualization of Spatial Distance.} We use generated frames (a) to reconstruct the point cloud (b), which is compared with the GT point cloud (c): Reconstructed point cloud (blue) is mostly overlapped with the GT point cloud, indicating our generated frames are spatially consistent. Distance between point clouds are used to measure the spatial consistency quantitatively.}
    \label{fig:evaluated_pc}
\end{figure}

\begin{figure}[t]
    \centering
    \includegraphics[width=\linewidth]{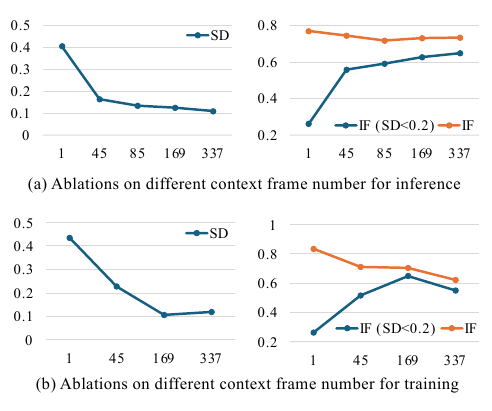}
    \vspace{-12pt}
    \caption{\textbf{Ablations on condition frame numbers for training and inference.} In the inference ablation (a), we fix the training number at 169; in the training ablation (b), we fix the inference number at 337.}
    \vspace{-18pt}
    \label{fig:ablation_chart}
\end{figure}

\begin{figure*}[t]
    \centering
    \includegraphics[width=0.99\linewidth]{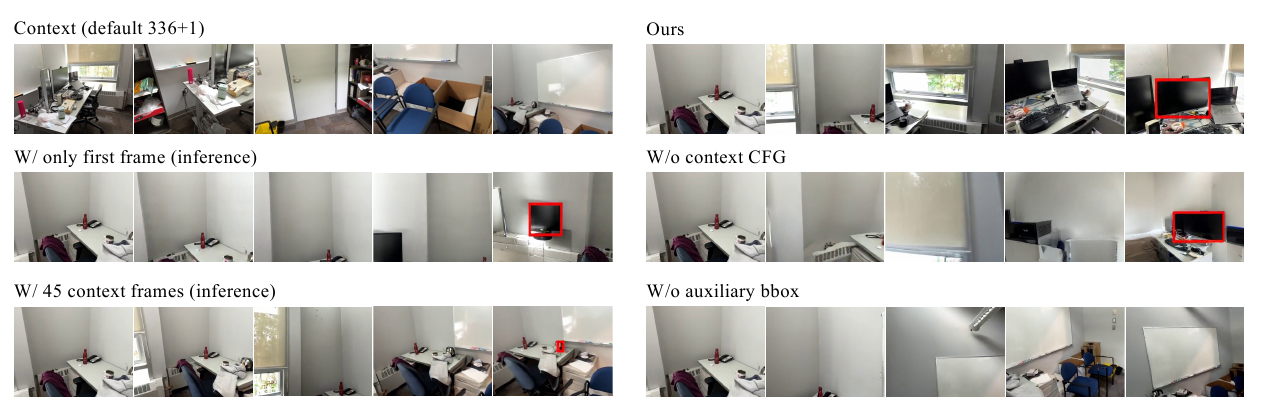}
    \vspace{-8pt}
    \caption{\textbf{Visualization for ablations on condition frame number, CFG, and auxiliary bbox.} The instruction is ``The camera moves through an office space. Finally, it focuses on \textbf{a monitor} in the center of the frame." For the best experience, see the supplementary videos.}
    \label{fig:ablation_fig}
    \vspace{-8pt}
\end{figure*}

\subsection{Video-based Object Grounding}\label{sec:eval_grounding}

Prior spatial-consistency metrics typically focus on static‑background consistency~\cite{huang2024vbench} or on the geometric consistency of generated videos~\cite{asim2025met3r}; and to our knowledge, there are no simliar setting on object grounding as ours. Accordingly, we propose two metrics tailored to this task.
\myparagraph{Spatial Distance (SD).} SD measures whether the generated scene is contained in the ground-truth(GT) point cloud and is an effective measure against out-of-context hallucination. We use VGGT~\cite{wang2025vggt} to reconstruct a point cloud \(\mathcal{P}^{\text{gen}}=\{\mathcal{P}^{\text{gen}}_i\}_{i\in\mathcal{I}}\) from each generated video, where $\mathcal{P}^{\text{gen}}_i$ is the point cloud of frame $i \in \mathcal{I}$. We calculate the maximum per-frame one-sided Chamfer distance of \(\mathcal{P}^{\text{gen}}\) to the ground-truth point cloud \(\mathcal{P}^{\text{gt}}\) as:
\begin{equation}
d\!\left(\mathcal{P}^{\text{gen}}, \mathcal{P}^{\text{gt}}\right)
= \max_{i \in \mathcal{I}}
\frac{1}{\lvert \mathcal{P}^{\text{gen}}_i \rvert}
\sum_{x \in \mathcal{P}^{\text{gen}}_i}
\min_{y \in \mathcal{P}^{\text{gt}}} \,\lVert x - y \rVert_2,
\end{equation}
and we report the mean of \(d\) over the test set. We use the maximum to penalize severe single‑frame geometric inconsistencies, which we regard as critical failures for navigation.
Figure~\ref{fig:evaluated_pc} shows that the point cloud reconstructed from VGGT using generated frames from our method align well with the ground-truth point cloud.

\myparagraph{Instruction Following (IF).}
To measure if the model is capable of navigating and locating the target object, we define IF score: for a total of $N$ generated videos, we use Qwen3-VL to check if the last frame of each video \(k \in [N]\) contains an object matching the target category, and if so, we further prompt Qwen3-VL to output bbox parameters $(x_k,y_k,W_k,H_k)$ for that object.
We consider the grounding successful if 
\[
\lvert x_k - x^{*} \rvert \le \tfrac{\gamma}{2}\, W_k
\quad\text{and}\quad
\lvert y_k - y^{*} \rvert \le \tfrac{\gamma}{2}\, H_k,
\]
where \((x^{*},y^{*})\) is the center of the frame, and \(\gamma\) is a tolerance ratio we set to 0.6.
The IF score is then defined to be the success rate of grounding across $N$ videos.


Since there might be scenarios where the object gets hallucinated in close to the final frames, achieving high IF score becomes meaningless if SD is high.
To evaluate spatial consistency and grounding jointly, we also propose SD-thresholded IF, denoted $\mathrm{IF}(\mathrm{SD} < \delta)$, which we calculate among videos of which SD is less than $\delta$. By default we evaluate using $\delta=0.2$ and $\delta=0.1$

Additionally, we report VBench Imaging Quality (IQ) and Dynamic Degree (DD)~\cite{huang2024vbench} to assess frame‑wise quality and motion dynamics as complementary metrics.

We evaluate object grounding on 107 prompts from the 18 validation scene in the form of: ``\texttt{The camera moves across <SCENE DESCRIPTION>. Finally, it focuses on <OBJECT DESCRIPTION> in the center of the frame.}" Table~\ref{tab:grounding_eval} reports results for our method alongside Wan2.2‑5B~\cite{wan2025wan}, Veo3~\cite{veo3} as SOTA open and closed-source image‑to‑video model, and FramePack~\cite{zhang2025packing} due to its capability to handle long context input.
Since those methods use different video‑conditioned settings, generation resolutions and the numbers of generated frames, we enumerate configuration differences in Table~\ref{tab:grounding_eval} for reference.  Notably, Wan2.2‑5B~\cite{wan2025wan} and Veo3~\cite{veo3} only support single image input; thus, the context frame count is 1. For FramePack~\cite{zhang2025packing}, we adopt its default setting of 105 context frames. Our default setting uses 337 context frames (including the first frame of the target video) of the generated video as context frames. By default, we generate 161 frames; when generating auxiliary bounding boxes at the video end, we add 20 frames (161+20). For reference, we also report the same metrics on ground‑truth (GT) videos for these 18 scenes.

We present quantitative results in Table~\ref{tab:grounding_eval} and qualitative examples in Fig.~\ref{fig:qualitative}. Our evaluation shows that though Wan2.2~\cite{wan2025wan} and Veo3~\cite{veo3} exhibit strong instruction following capability, their generation is often hallucinatory: without video conditioning, their SD and IF(SD$<\delta$) degrade markedly. FramePack~\cite{zhang2025packing} conditions on history frames, yet still hallucinates due to heavily compressed context tokens that discard critical spatial cues. In contrast, our method delivers faithful grounding while preserving scene consistency: our SD approaches that of GT, indicating strong geometric alignment, and we achieve competitive IQ alongside high motion dynamics on DD.

\begin{figure*}[t]
    \centering
    \includegraphics[width=\linewidth]{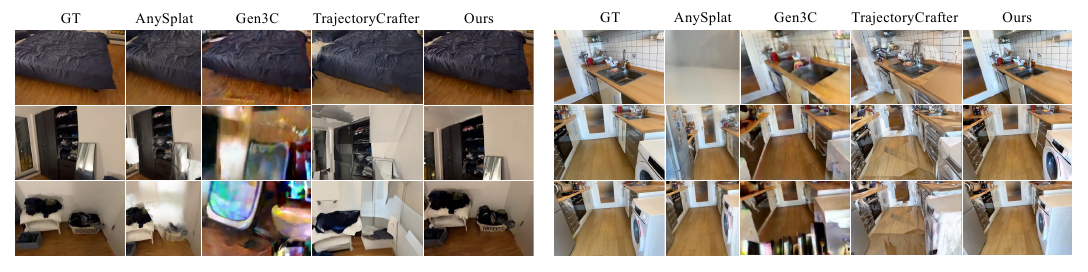}
    \vspace{-12pt}
    \caption{\textbf{Qualitative results on scene navigation of a bedroom(left) and a kitchen(right).} Our method delivers the highest perceptual quality and good camera controllability. Please refer to the supplementary videos for clearer visual comparisons.}
    \vspace{-12pt}
    \label{fig:vis_camctrl}
\end{figure*}

\subsection{Ablations on Object Grounding}

We further ablate several key design choices. The main results are in in the bottom part of Table~\ref{tab:grounding_eval} and Fig.~\ref{fig:ablation_chart} and ~\ref{fig:ablation_fig}.

\myparagraph{Context frame length.} 
Context length is a key factor for both training and inference. Our default uses 169 frames (43 latent frames) for training and 337 frames (85 latent frames) for inference. As shown in Table~\ref{tab:grounding_eval}, using only the first frame without other context yields a much higher SD (0.4043) than using context (0.1099), indicating substantially greater inconsistency relative to the contextual environment. Conversely, IF improves with only the first frame because generation is no longer constrained and may hallucinate the target object for grounding.

To determine the optimal context length, we conduct ablations study shown in Fig.~\ref{fig:ablation_chart}. In (a), we train with 169 context frames and vary inference from 1–337 frames. Both SD and IF (SD$<0.2$) improves monotonically as the context frame number increases. Interestingly, IF first decreases, then rises, and finally saturates. With insufficient context, the model tends to hallucinate, producing poor spatial consistency (higher SD) but faked grounding (higher IF). As more context frames are provided, generation becomes constrained; IF initially drops when the model is constrained yet lacks sufficient evidence for grounding, then increases once context is rich enough to support accurate scene understanding.

In (b), we vary the training context from 1–337 frames while fixing inference at 337 frames. All models share identical settings and sampling steps, differing only in training context frame length. Increasing training context generally improves spatial consistency (lower SD), whereas IF (SD$<0.2$) first increases, then declines, with a pronounced drop at 337 frames. We hypothesize that excessive training context biases the model toward adhering to visual context at the expense of following text instructions. Training with 169 context frames provides a strong balance.

Notably, models trained with shorter context frames (\eg, 169) generalize well to longer inference context frames (\eg, 337).

\myparagraph{Auxiliary bounding box.} Auxiliary bbox design plays a significant role in improving grounding accuracy. As shown in Table~\ref{tab:grounding_eval}, training the model to predict the target object’s bbox raises IF(SD$<0.2$) from 0.5401 to 0.6486. In Fig.~\ref{fig:ablation_fig} we show example output with and without auxiliary bbox: for the “monitor” target, prediction without the bbox leads the model to drift to irrelevant objects, whereas providing the bbox performs correct grounding of the monitor.

\myparagraph{CFG.} As shown in Table~\ref{tab:grounding_eval}, with no CFG (pure conditional generation), both SD and IF(SD$<\delta$) degrade markedly. If we retain text CFG but remove context CFG, the model tends to hallucinate without enforcing contextual consistency: IF appears high, but SD and IF(SD$<\delta$) are significantly worse. 


\myparagraph{Non‑continuous RoPE for non‑contiguous context.} We find that non‑continuous RoPE helps the model reason over non‑contiguous context. As shown in Table~\ref{tab:grounding_eval}, using vanilla (continuous) RoPE over non‑contiguous context substantially worsens SD and IF (SD$<\delta$), indicating poorer spatial understanding of the contextual environment.

\myparagraph{Pretraining.}
Pretraining confers clear benefits; without it (\ie, training from scratch), SD, IF, IF(SD$<\delta$), and IQ degrade markedly (Fig.~\ref{tab:grounding_eval}).

\subsection{Video-based Scene Navigation}\label{sec:eval_scene_nav}
For scene navigation, we report PSNR, LPIPS and IQ for generation fidelity and adherence to the requested camera trajectory.

We evaluate video-based scene navigation on 18 ScanNet++~\cite{yeshwanth2023scannet++} scenes, each with 5 camera trajectories (90 cases in total). By default, we condition on 337 context frames (85 latent) and generate 161 output frames (41 latent). We compare against the feed-forward 3D reconstruction method AnySplat~\cite{jiang2025anysplat} and the video generation model Gen3C~\cite{ren2025gen3c} and TrajectoryCrafter~\cite{yu2025trajectorycrafter}. Notably, all these methods rely on external estimators (\eg, VGGT~\cite{wang2025vggt}) for extra 3D information.
For our method, it only take context frames and no 3D information is needed. See the supplementary material for detailed settings.

In Table~\ref{tab:nvs_eval}, we report comparative results. Because PSNR and LPIPS emphasize structural similarity, AnySplat~\cite{jiang2025anysplat} attains the highest scores on these metrics, though with significant artifacts it acheive lower perceptual IQ score. Our method delivers competitive PSNR and LPIPS while achieving substantially better IQ. Moreover, we outperform the video‑based Gen3C~\cite{ren2025gen3c} and TrajectoryCrafter~\cite{yu2025trajectorycrafter} across all metrics; unlike these video-based methods, which heavily depends on external estimators, our approach achieves strong end‑to‑end performance. Fig.~\ref{fig:vis_camctrl} further visualizes the results: although AnySplat achieves high PSNR, it exhibits noticeable artifacts, whereas our outputs are visually cleaner but not perfectly aligned with GT. We futher supply qualitative results for visualization in Fig.~\ref{fig:vis_camctrl}

\begin{table}[t]
\centering
\caption{\textbf{Evaluation on video‑conditioned scene navigation.} “3D info.” indicates explicit 3D information required for each method. Our method does not need any explicit 3D information.}

\vspace{-6pt}
\label{tab:nvs_eval}
\small
\resizebox{0.48\textwidth}{!}{
\begin{tabular}{l cccc}
\toprule
Method  & 3D info. & PSNR $\uparrow$ & LPIPS $\downarrow$ & IQ $\uparrow$ \\
\midrule
\multicolumn{3}{l}{3D reconstruction} \\
\midrule
AnySplat~\cite{jiang2025anysplat} & Cam. pose & \textbf{16.40} & \textbf{0.4402} & 0.4568 \\
\midrule
\multicolumn{3}{l}{Video generation} \\
\midrule
TrajectoryCrafter~\cite{yu2025trajectorycrafter} & Cam. pose, depth & 13.56 & 0.4920 & \underline{0.5630} \\
Gen3C~\cite{ren2025gen3c} &  Cam. pose, depth & 12.76 & 0.5695 & 0.3628 \\
Ours & - & \underline{14.27} & \underline{0.4674} & \textbf{0.6320} \\
\bottomrule
\end{tabular}
}
\vspace{-16pt}
\end{table}

\subsection{Out-of-domain generalization.}

\begin{figure}[t]
    \centering
    \includegraphics[width=\linewidth]{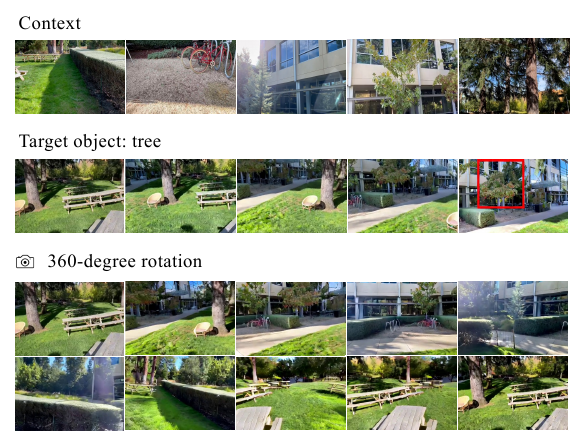}
    \vspace{-18pt}
    \caption{\textbf{Out-of-domain result.} Our models can generalize to OOD scenarios like outdoor scenes and new categories, performing object grounding and scene navigation.}
    \label{fig:ood}
\end{figure}

Although trained on only indoor scenes, our model generalizes well to out-of-domain environments for both tasks. In Fig.~\ref{fig:ood}, we show results on a real-captured outdoor park: the model reliably grounds object categories that were rarely encountered during indoor training (\eg, trees), and supports free-form navigation, including 360° rotations.
\section{Conclusions and Future Work}

We introduced \nickname, a video-only framework that probes visuospatial intelligence in generative models by conditioning solely on scene context. Across scene navigation and object grounding, \nickname executes geometry-consistent camera trajectories and good target grounding, while generalizing to long contexts and out-of-domain environments.

Looking ahead, we note several limitations and avenues for future work. Our current approach operates at a modest $416 \times 256$ resolution due to the absence of context compression (\eg, \cite{zhang2025test, gu2024mamba}), which we identify as a key lever for improving visual fidelity at higher resolutions. In addition, stronger temporal modeling, targeted data augmentation, and improved grounding objectives could mitigate artifacts such as temporal discontinuities and incorrect grounding on long‑tail categories. Finally, extending this framework beyond static scenes to complex, dynamic environments and a broader range of tasks is an exciting direction.


\clearpage

\section{Suppplementary Materials}

\subsection{More Qualitative Results}
It is highly recommended to refer to the webpage for visualization.

\subsection{Implementation Details.}

\myparagraph{Data Curation.}
We sample frames from ScanNet++~\cite{yeshwanth2023scannet++} and ARKitScenes~\cite{baruch2021arkitscenes} at an interval of three to reduce temporal redundancy. To curate datasets for object grounding, for each sampled frame, we use Qwen3‑VL~\cite{qwen3vl} to perform object grounding with the following prompt template:
\begin{tcolorbox}[enhanced,breakable,
colback=white,colframe=black,
title=Object Grounding Prompt Template,
colbacktitle=black!85,coltitle=white,fonttitle=\bfseries,
boxed title style={sharp corners,boxrule=0pt},
attach boxed title to top left={yshift=-2mm}
]
You are a helpful assistant. Identify ALL objects that belong to these categories: \texttt{YOUR\_CATEGORY}.
Requirements:
\begin{enumerate}[leftmargin=3pt,itemsep=3pt]
\item Return ALL instances of objects from these categories (can be multiple).
\item Each object must be CLEARLY VISIBLE with SHARP, DISTINCT boundaries (not blurry or pixelated).
\item If the image is blurry, low quality, or no clear objects exist, return an empty array: [].
\item Format: [{"label": "DETECTED\_CATEGORY", "box": [x1, y1, x2, y2]}, ...]
\end{enumerate}
\end{tcolorbox}
We then filter detections using a center‑region threshold (center ratio) and select the instance closest to the image center as the grounding target. For each clip, we take a frame where the target is centered as the final frame and extract the preceding 161 frames (inclusive) to form the target video clip. We also use Qwen3‑VL to caption the clip with the following prompt:
\begin{tcolorbox}[enhanced,breakable,
colback=white,colframe=black,
title=Video Caption Prompt Template,
colbacktitle=black!85,coltitle=white,fonttitle=\bfseries,
boxed title style={sharp corners,boxrule=0pt},
attach boxed title to top left={yshift=-2mm}
]
You are a helpful assistant. Describe a video sequence in which the camera moves through the environment and, at the end, a \{centered\_object\_label\} appears centered in the frame. Use simple sentences. Do NOT use complex grammar.
Good example: ``The camera moves through a room. It pans left and right. At the end, a \{centered\_object\_label\} is centered in the frame.''
Bad example: ``The camera pans around, revealing various objects and eventually discovering a \{centered\_object\_label\} positioned in the center, where it comes to rest in the frame.''
Provide a concise description in 1–2 simple sentences.
\end{tcolorbox}

\myparagraph{Training with context.}
During training, we sample both context frames and target generation frames from source videos. Target clips are 161 frames long. Context frames are randomly drawn but always in contiguous groups of four to satisfy the Wan2.2~\cite{wan2025wan} VAE encoder. For efficiency, we pre‑encode frames into latents during preprocessing and sample latents directly, yielding roughly 2× faster training.
A key consideration is ensuring sufficient overlap between context and target frames so the model can learn geometric consistency. ScanNet++ sequences are long with ample overlap, so we sample context and targets from the same clip. ARKitScenes sequences are shorter with less view overlap but include multiple videos per environment; therefore, we sample context and targets from different videos within the same environment.

\myparagraph{Spatial Distance.}
To measure the distance between the point cloud reconstructed by VGGT~\cite{wang2025vggt} and the ground‑truth point cloud, we first register the two coordinate frames. ScanNet++~\cite{yeshwanth2023scannet++} provides frames with calibrated camera poses and depth maps. When reconstructing point clouds from generated videos, we append 40 ground‑truth frames as anchors and use the estimated point clouds of these anchors to perform registration. After alignment, we compute the point‑cloud distance.

\begin{figure*}[t]
    \centering
    \includegraphics[width=\linewidth]{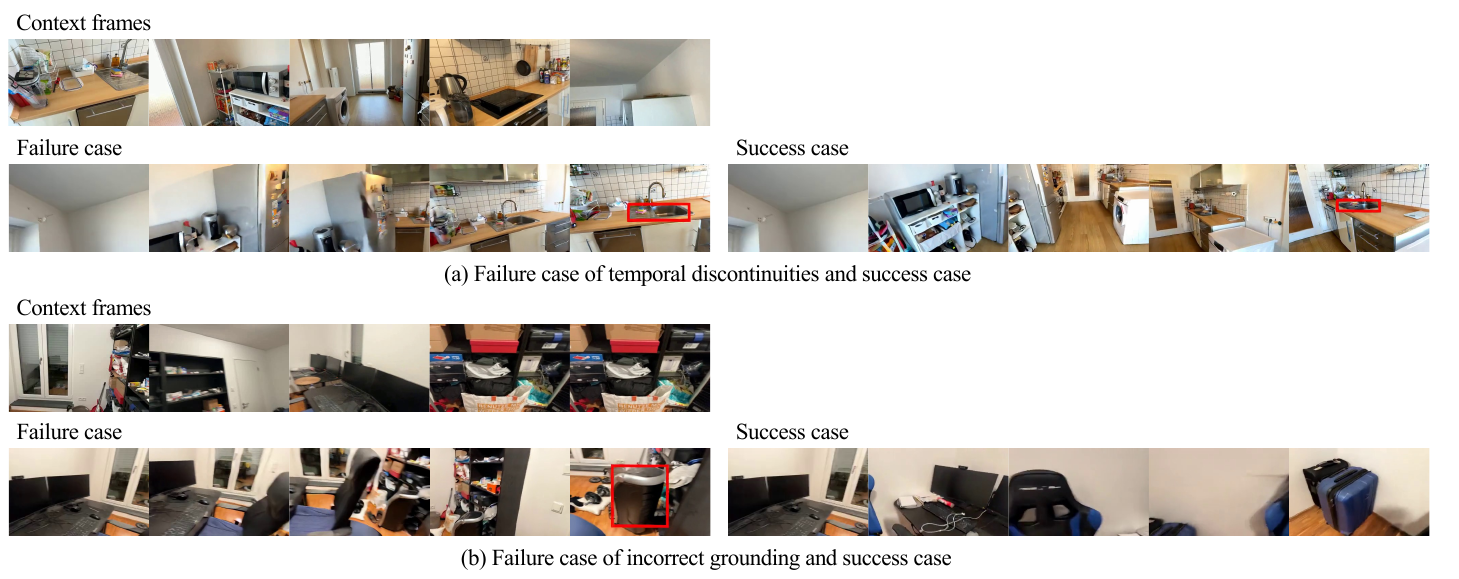}
    \caption{Failure and success cases.}
    \label{fig:failurecase}
\end{figure*}

\begin{figure}[t]
    \centering
    \includegraphics[width=0.5\linewidth]{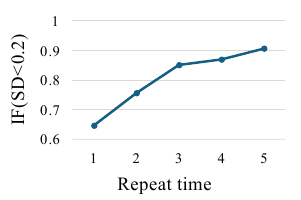}
    \caption{Ablation on grounding success rate over repeat time.}
    \label{fig:repeattime}
\end{figure}

\begin{figure}[t]
    \centering
    \includegraphics[width=\linewidth]{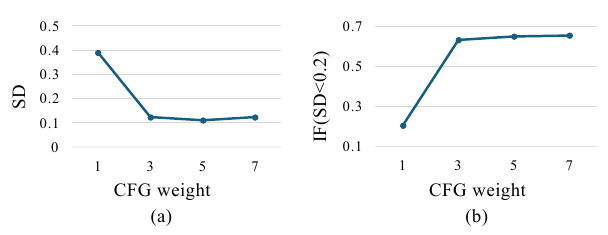}
    \caption{Ablation on different CFG weights  .}
    \label{fig:cfg_weight_ablation}
\end{figure}

\subsection{Benchmark setting.}
Since it is not feasible to make the setting for different methods perfectly fair, we list the detailed setting for each methods for reference.

 \myparagraph{Wan2.2-5B~\cite{wan2025wan}.} We generate 121-frame videos at a resolution of $1280 \times 704$. The model takes the first frame and a text instruction as input, utilizing 50 inference steps and a classifier-free guidance scale of 5.0.

\myparagraph{Veo 3~\cite{veo3}.} We utilize the Veo 3 image to video model via its official API, generating videos at $1280 \times 720$ with 192 frames.

\myparagraph{FramePack~\cite{zhang2025packing}.} We employ FramePack to generate videos at a resolution of $704 \times 544$. The inference process utilizes 50 sampling steps with a classifier-free guidance scale of 3.0. The model is conditioned on a context window of 105 frames, which are encoded into a multi-scale latent representation comprising 16 latent frames at 4x compression, 2 latent frames at 2x compression, and 9 latent frames at 1x compression.

\myparagraph{Anysplat~\cite{jiang2025anysplat}.} We employ Anysplat for 3D reconstruction and rendering for video outputs. The model utilizes 84 context frames along with the first frame serving as an anchor for reconstruction, which correspond to 84 latent frames plus the first frame for the video generation setting. We estimate camera poses using VGGT~\cite{wang2025vggt} and use them for coordinate regularization.

\myparagraph{Gen3C~\cite{ren2025gen3c}.} We follow the official implementation of Gen3C and its instruction on Multiview Images Input. First, we run VGGT ~\cite{wang2025vggt} to get the depth information, camera intrinsics, and extrinsics of the context frame. We also include the first frame of the generation as the first context frame. During the VGGT inference, we include context frames and ground truth frames as a single run to get the camera poses of the ground truth. During the Gen3C inference, we use the camera poses of ground truth as camera control. We input 85 context frames and output 121 frames, all at resolution 576x320.

\myparagraph{TrajectoryCrafter~\cite{yu2025trajectorycrafter}.} We use the same preprocessing algorithm as Gen3C, \ie VGGT, to get the point cloud from 85 context frames and camera pose control from 121 ground truth frames, all in one pass to make sure they are in the same corodinate system. Then we project the point cloud based on ground truth camera and get masks. Since the model is trained to generate 49 frames, we run three inferences to get the complete output.

\subsection{Experiments.}

\myparagraph{Repeat time.} Consistent with prior findings that Veo3~\cite{veo3} improves with more repeats~\cite{wiedemer2025video}, we observe the same trend. For object grounding, we compute IF (SD$<0.2$) across repeated runs and deem a case successful if any repeat succeeds. As shown in Fig.~\ref{fig:repeattime}, increasing the number of repeats from 1 to 5 raises IF (SD$<0.2$), indicating that repeated sampling substantially boosts performance.

\myparagraph{CFG strength.} We investigate the impact of different CFG scales in Fig.~\ref{fig:cfg_weight_ablation}. While a weight of 1 (no CFG) yields suboptimal results, performance improves significantly once the scale reaches a sufficient level (e.g., 3) and remains robust across a wide range (e.g., 3--7). This suggests that the use of CFG itself is more critical than fine-tuning the exact weight.

 \myparagraph{Non-continuous context sampling.} We investigate the effectiveness of non-continuous context sampling during training, as shown in Table~\ref{tab:non_contunuous_context_sampling}. Our results demonstrate that non-continuous sampling significantly improves geometric consistency with the context and achieves highly faithful grounding. We hypothesize that this improvement stems from the model developing a more robust spatial understanding through such diverse and non-continuous context modeling.

\begin{table}[t]
\centering
\caption{Ablation on context sampling.}
\label{tab:non_contunuous_context_sampling}
\small
\resizebox{0.43\textwidth}{!}{
\begin{tabular}{l ccc}
\toprule
  & SD$\downarrow$ & IF$\uparrow$ & IF(SD$<0.2$)$\uparrow$ \\
\midrule
Non-continuous & 0.1099 & 0.7327 & 0.6486 \\
Continuous & 0.3246 & 0.8497 & 0.4600 \\
\bottomrule
\end{tabular}
}
\end{table}

\myparagraph{Random zero camera pose.} Existing camera control methods typically normalize the camera pose of the first frame to zero during both training and inference. We find that, particularly with limited datasets, fixing the first frame to the origin restricts spatial exploration, thereby limiting generalization in complex scenarios such as $360^\circ$ rotations. To address this, we propose randomly selecting a reference frame to serve as the zero pose, which effectively alleviates this issue and improves generalization.


 \myparagraph{Computational analysis.} Although our model processes relatively long contexts (default 337 frames) and generates extended sequences (default 181 frames), we maintain computational costs within an acceptable range. Inference requires approximately 34GB of VRAM and takes 2 minutes on a single A100 GPU using CFG and 50 denoising steps, without optimizations such as VAE slicing, tiling, or dynamic model loading~\cite{zhang2025packing}. This efficiency stems primarily from two factors: (1) the use of a moderate resolution ($416\times 256$), which we find sufficient for high visual quality and experimental validation; and (2) the Wan2.2 VAE, which achieves high spatial and temporal compression ratios while preserving visual fidelity.

\myparagraph{Failure cases.}
Our method still suffers from artifacts such as temporal discontinuities (Fig.~\ref{fig:failurecase}(a)) and incorrect grounding (Fig.~\ref{fig:failurecase}(b)) for some cases.

\clearpage

{
    \small
    \bibliographystyle{ieeenat_fullname}
    \bibliography{main}
}


\end{document}